\documentclass{article}

    \usepackage[preprint]{neurips_2025}

\usepackage[utf8]{inputenc} 
\usepackage[T1]{fontenc}    
\usepackage{hyperref}       
\usepackage{url}            
\usepackage{booktabs}       
\usepackage{amsfonts}       
\usepackage{nicefrac}       
\usepackage{microtype}      
\usepackage{xcolor}         

\usepackage{graphicx}
\usepackage{booktabs}  
\usepackage{multirow}  
\usepackage{lineno}
\usepackage{subcaption}
\usepackage{epigraph}
\usepackage{amsmath}
\usepackage{amssymb}

\definecolor{darkblue}{rgb}{0, 0, 0.5}

\title{\texttt{Min-p}, Max Exaggeration: A Critical Analysis of\\\texttt{Min-p} Sampling in Language Models}

\author{%
  Rylan Schaeffer\\
  Stanford University\\
  \texttt{rschaef@cs.stanford.edu}\\
  \And
  Joshua Kazdan\\
  Stanford University \\
  \texttt{jkazdan@stanford.edu} \\
  \And
  Yegor Denisov-Blanch\\
  Stanford University \\
  \texttt{ydebl@stanford.edu} \\
}

\begin{document}

\maketitle

\begin{abstract}
Sampling from language models impacts the quality and diversity of generated outputs, affecting both research and real-world applications.
Recently, \citet{nguyen2024minp}'s ``Turning Up the Heat: Min-p Sampling for Creative and Coherent LLM Outputs'' introduced a new sampler called \texttt{min-p}, claiming it achieves superior quality and diversity over established methods such as \texttt{basic}, \texttt{top-k}, and \texttt{top-p} sampling.
The significance of these claims was underscored by the paper's recognition as the 18th highest-scoring submission to ICLR 2025 and selection for an Oral presentation.
This paper conducts a comprehensive re-examination of the evidence supporting \texttt{min-p} and reaches different conclusions across the original paper's four main lines of evidence.
First, the original paper's human evaluations omitted data, conducted statistical tests incorrectly, and described qualitative feedback inaccurately; our reanalysis demonstrates \texttt{min-p} did not outperform baseline samplers in quality, diversity, or a trade-off between quality and diversity. In response to our findings, the authors of the original paper conducted a new human evaluation using a different sampler implementation, task, and rubric that nevertheless provides further evidence \texttt{min-p} does not improve over baseline samplers.
Second, comprehensively sweeping the original paper's NLP benchmark evaluations reveals \texttt{min-p} does not surpass baseline samplers when controlling for the number of hyperparameters.
Third, the original paper's LLM-as-a-Judge evaluations lack methodological clarity and appear inconsistently reported: the higher of two scores was reported for \texttt{min-p} while the lower of two scores was reported for \texttt{top-p}.
Fourth, community adoption claims (49k GitHub repositories, 1.1M GitHub stars) were found to be unsubstantiated, leading to their removal from the ICLR 2025 Camera Ready; the revised adoption claim remains misleading.
We conclude that evidence presented in the original paper fails to support claims that \texttt{min-p} improves quality, diversity, or a trade-off between quality and diversity.
\end{abstract}

\section{Introduction}

Large language model (LLM) capabilities have transformed numerous domains, from creative writing to scientific research. A critical detail of LLM deployment is the \emph{sampling method}: the algorithm that determines how tokens are sampled during generation.
Sampling strategies directly impact the quality and diversity of generated outputs, making them important to both research and deployment.

Commonly used samplers include \texttt{basic} (temperature-only) \texttt{sampling} \citep{ackley1985learning}, which samples tokens based on their temperature-scaled softmax-normalized logits; \texttt{top-k sampling} \citep{fan2018topk}, which samples the $k$ most probable tokens; and \texttt{top-p sampling} \citep{holtzman2020topp}, which samples tokens comprising the top $p$ probability mass. Other samplers include \texttt{$\eta$-sampling}, $\texttt{$\epsilon$-sampling}$ \citep{hewitt2022truncation} and \texttt{mirostat sampling} \citep{basu2020mirostat}.

Recently, the paper ``Turning Up the Heat: Min-P Sampling for Creative and Coherent LLM Outputs" \citep{nguyen2024minp} introduced a new sampling method called \texttt{min-p sampling}, claiming it produces higher quality and higher diversity outputs than other samplers.
Given the potential impact of an improved sampling method and the paper's exposure as the 18th highest-scoring submission at ICLR 2025\footnote{\url{https://papercopilot.com/statistics/iclr-statistics/iclr-2025-statistics/}}, we carefully scrutinized the methodologies, data, analyses, code and conclusions presented in support of \texttt{min-p} across the authors' four lines of evidence: (1) human evaluations, (2) natural language processing (NLP) benchmark evaluations, (3) LLM-As-A-Judge evaluations and (4) community adoption metrics.
Our re-analyses of the evidence lead us to conclude that \textbf{relative to commonly used samplers, \texttt{min-p} does not improve quality or diversity or the trade-off between quality and diversity}.
Our code is \href{https://github.com/RylanSchaeffer/KoyejoLab-Min-p-Sampling}{publicly available on GitHub}, as are \href{https://wandb.ai/rylan/min-p-evals/sweeps}{our W\&B sweeps} of NLP benchmark evaluations.

\section{Re-Analyzing \texttt{Min-p}'s Human Evaluations}
\label{sec:human_evals}

We began with re-analyzing the original paper's human evaluations since human judgments are widely considered the gold standard for assessing language model outputs \citep{van2019best,roller2020opendomainconversationalagentscurrent,howcroft2020twenty,clark2021all,liang2022holistic,khashabi2022geniereproduciblestandardizedhuman,chiang2024chatbotarenaopenplatform,biderman2024lessonstrenchesreproducibleevaluation,schaeffer2025correlatingpredictinghumanevaluations}. 
We identified four key issues.

\subsection{Human evaluators’ scores for one of two baseline samplers were omitted}
\label{sec:human_evals:subsec:omitted_data}

Section 6 of \citet{nguyen2024minp} states human participants evaluated \texttt{min-p} against a single sampler: \texttt{top-p}.
Both the \href{https://arxiv.org/abs/2407.01082v2}{Oct 2024 Arxiv manuscript} and \href{https://openreview.net/notes/edits/attachment?id=SR6ORZeBAn&name=pdf}{ICLR OpenReview manuscript} repeatedly state that \texttt{min-p} and \texttt{top-p} were considered, and their Table 4 presents results only for these.
However, when examining the \href{https://github.com/menhguin/minp_paper/blob/main/min_p%20user%20preference%20study%20v2.0%20(Responses)%20-%20Form%20responses%201%20(1)%20(2).csv}{paper's data}, we discovered that \textbf{scores for a second baseline sampler (\texttt{basic} sampling) were excluded from the methodology, the analysis and the results without mention or explanation}. 
We \href{https://github.com/menhguin/minp_paper/issues/4}{publicly confirmed with the authors}.
These omitted scores comprised $1/3^{\text{rd}}$ of the total collected scores.
After we raised the issue, the omitted data were added to the \href{https://openreview.net/notes/edits/attachment?id=oQLYV9wuaa&name=pdf}{Camera Ready's Table 4}, but the methodology, the results and the conclusions have not been correspondingly updated.

\begin{figure}[t!]
    \centering
    \begin{minipage}{\textwidth}
        \centering
        \includegraphics[width=\linewidth]{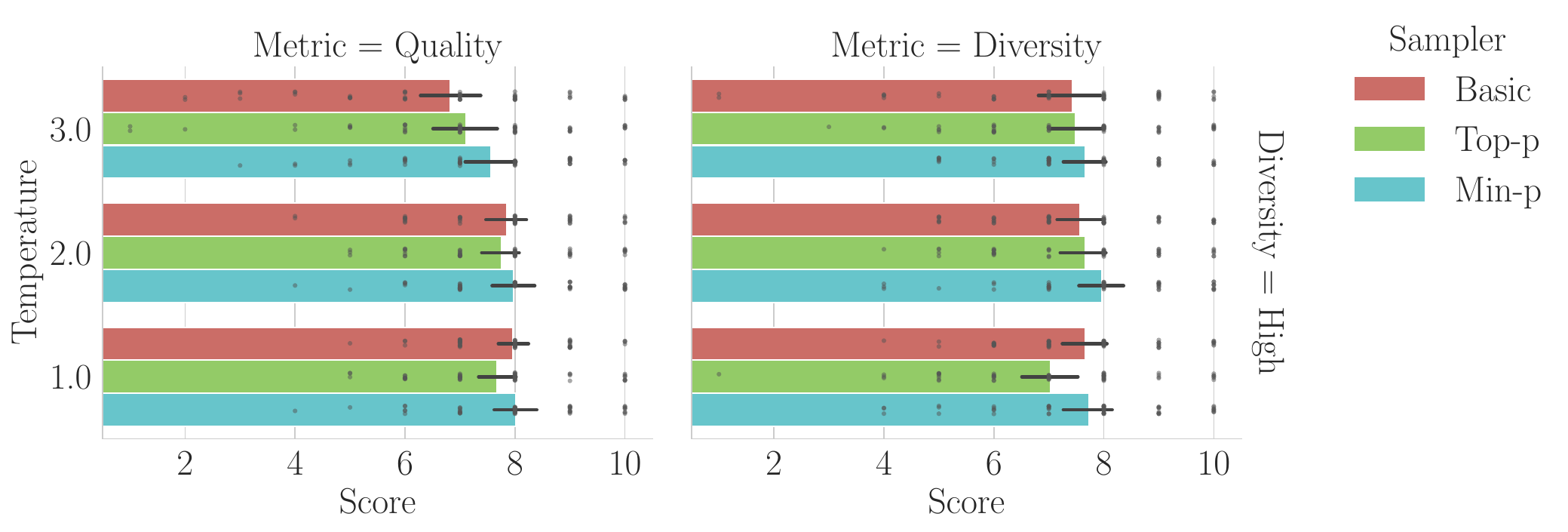}
        \captionof{figure}{\textbf{Visualizing Human Evaluators' Scores from \citet{nguyen2024minp}'s Data Demonstrates \texttt{Min-p} Does Not ``Consistently" Outperform Other Samplers.} Rather, the original paper's data suggest \texttt{min-p} is largely indistinguishable from other samplers based on 95\% confidence intervals.}
        \label{fig:human_evals}
    \end{minipage}
\end{figure}

\subsection{Visualizations and Statistical Tests Fail to Support Claim That \texttt{Min-p} Outperforms Other Samplers}
\label{sec:human_evals:subsec:human_scores}

Based on the human evaluators' scores, Section 6 of \citet{nguyen2024minp} concluded that \texttt{min-p} ``consistently" outperformed \texttt{top-p} ``across all settings":
\begin{quote}
    ``Overall, \texttt{min-p} sampling consistently scored higher than \texttt{top-p} sampling across all settings [...] A paired t-test confirmed that the differences in scores between min-p and top-p sampling were statistically significant ($p < 0.05$).''
\end{quote}
However, \textbf{both visualizations and statistical hypothesis tests of the original human evaluation data suggest \texttt{min-p} is indistinguishable from the baselines in almost all settings.}

To briefly explain the human evaluation methodology, three samplers (\texttt{basic}, \texttt{top-p} and \texttt{min-p}) were compared in six conditions: three temperatures ($1.0, 2.0, 3.0$) and two diversity settings (``high" and ``low") corresponding to different $p$ hyperparameters.
Humans were asked to score the generated outputs under two metrics: quality and diversity.
Participants were excluded if they failed attention checks.
For more information, please see the original manuscript.

\begin{table}[t!]
    \centering
    \begin{tabular}{llcccccc}
    \toprule
    \multirow{2}{*}{Metric} & \multirow{2}{*}{Alt. Hyp.} & \multicolumn{6}{c}{Temperature} \\
    \cmidrule(lr){3-8}
    & & \multicolumn{2}{c}{$\tau=1.0$} & \multicolumn{2}{c}{$\tau=2.0$} & \multicolumn{2}{c}{$\tau=3.0$} \\
    \cmidrule(lr){3-4} \cmidrule(lr){5-6} \cmidrule(lr){7-8}
    & & $t$ & $p$ & $t$ & $p$ & $t$ & $p$ \\
    \midrule
    \multirow{2}{*}{Quality} & \texttt{Min-p} $>$ \texttt{Basic} & 0.33 & .370 & 0.65 & .260 & 3.13$^{***\dagger}$ & .001 \\
    & \texttt{Min-p} $>$ \texttt{Top-p} & 2.05$^{*}$ & .023 & 1.18 & .121 & 2.02$^{*}$ & .025 \\
    \midrule
    \multirow{2}{*}{Diversity} & \texttt{Min-p} $>$ \texttt{Basic} & 0.31 & .378 & 1.86$^{*}$ & .034 & 0.85 & .201 \\
    & \texttt{Min-p} $>$ \texttt{Top-p} & 2.64$^{**}$ & .006 & 1.44 & .078 & 0.87 & .195 \\
    \bottomrule
    \multicolumn{8}{l}{\scriptsize{$^{*}$ $p < 0.05$, $^{**}$ $p < 0.01$, $^{***}$ $p < 0.001$, $^{\dagger}$ Significant after Bonferroni correction for 12 comparisons.}} \\
    \multicolumn{8}{l}{\scriptsize{Note: All tests were paired t-tests with df = 52, one-sided (alternative = "greater")}} \\
    \phantom{Blank Line}\\
    \end{tabular}\\
    \captionof{table}{\textbf{Hypothesis Testing of Human Evaluators' Scores Fails to Support Claim that \texttt{Min-p} Consistently Outperforms Other Samplers.} To test whether evidence supports the claim that \texttt{min-p} ``consistently outperforms" other samplers, we conducted one-sided paired t-tests using the authors' published data. Without correcting for multiple comparisons, evidence exists to support \texttt{min-p}'s superiority in 5 of 12 comparisons at $\alpha=0.05$ and 2 of 12 comparisons at $\alpha=0.01$. After applying a Bonferroni correcting for multiple comparisons, evidence exists to support \texttt{min-p}'s superiority in 1 of 12 comparisons at $\alpha=0.05$ and 0 of 12 comparisons at $\alpha=0.01$. For details, see Sec.~\ref{sec:human_evals:subsec:human_feedback}.}
    \label{tab:human_evals_paired_ttests}
\end{table}

We focused on the ``high" diversity setting for three reasons: 
First, the claimed advantage of \texttt{min-p} sampling is that it provides both high quality and high diversity, whereas other samplers typically trade one off against the other. Second, \href{https://github.com/menhguin/minp_paper/issues/4}{the authors publicly told us to focus on the high diversity setting}, writing that ``the low [diversity] settings were quite experimental". Third, we believe that \texttt{top-p}'s $p$ value in the low diversity setting was poorly chosen; indeed, after we raised these concerns, the authors ran a new human evaluation that changed the low diversity \texttt{top-p} $p$ from $0.1$ to $0.9$. We return to this second new human evaluation in Sec.~\ref{sec:human_evals:subset:new_exp}.

We began by visualizing the human evaluations' scores from \citet{nguyen2024minp}. \textbf{Using the original paper's data, Fig.~\ref{fig:human_evals} reveals that the three samplers provide similar quality and similar diversity, with 95\% confidence intervals frequently overlapping}.

To more rigorously assess the claim that \texttt{min-p} consistently outperforms other samplers, we conducted 12 one-sided paired t-tests for each metric (quality or diversity), temperature ($1.0, 2.0, 3.0$) and baseline sampler (\texttt{min-p} versus \texttt{basic}, \texttt{min-p} versus \texttt{top-p}). In each test, the null hypothesis is \texttt{min-p}'s score is less than or equal to the other sampler's score, and the alternative hypothesis is \texttt{min-p}'s score is greater than the other sampler's score.
Statistical test results are displayed in Table~\ref{tab:human_evals_paired_ttests}.
Without correcting for multiple comparisons, we found evidence to reject the null hypotheses in 5 of 12 tests at $\alpha=0.05$ and 2 of 12 tests at $\alpha=0.01$. After applying a Bonferroni correction for multiple comparisons, we found evidence to reject the null hypothesis in 1 of 12 tests at $\alpha=0.05$ and 0 of 12 tests at $\alpha=0.01$.
\textbf{Based on the original paper's data, there is insufficient evidence to support the claim that \texttt{min-p} consistently outperforms baseline samplers across all settings.}

Furthermore, given that the original paper claims that \texttt{min-p} ``consistently" scores higher, an Intersection-Union Test (IUT) may be the appropriate statistical test, where the alternative hypothesis is that \texttt{min-p} is better in all 12 comparisons and the null hypothesis is the set complement. Since the largest p-value of the 12 comparisons is $0.378$, under the IUT, we again find insufficient evidence to reject the null hypothesis at both $\alpha=0.05$ and $\alpha=0.01$.

The original paper's statistical analysis reached a different conclusion for two reasons. First, despite claiming that \texttt{min-p} ``consistently scored higher" ``across all settings"  (metric, temperature, and diversity), the paper pooled data across all settings and performed a single t-test, which tests whether \texttt{min-p} scored higher on average. Second, pooling over all settings is misleading in that in the ``low'' diversity condition, \texttt{top-p}'s hyperparameter $p$ was poorly chosen in a way that pulled \texttt{top-p} down significantly; the authors said publicly to ignore this particular low diversity condition and subsequently changed $p$ in their new human experiment (Sec.~\ref{sec:human_evals:subset:new_exp}).
Thus, we believe the original paper's statistical inferences are misleading or incorrect.

\subsection{Human Evaluators' Qualitative Responses Fail to Support Claim That \texttt{Min-p} Is Preferred Over Other Samplers}
\label{sec:human_evals:subsec:human_feedback}

\begin{figure}[t!]
    \centering
    \includegraphics[width=0.7\linewidth]{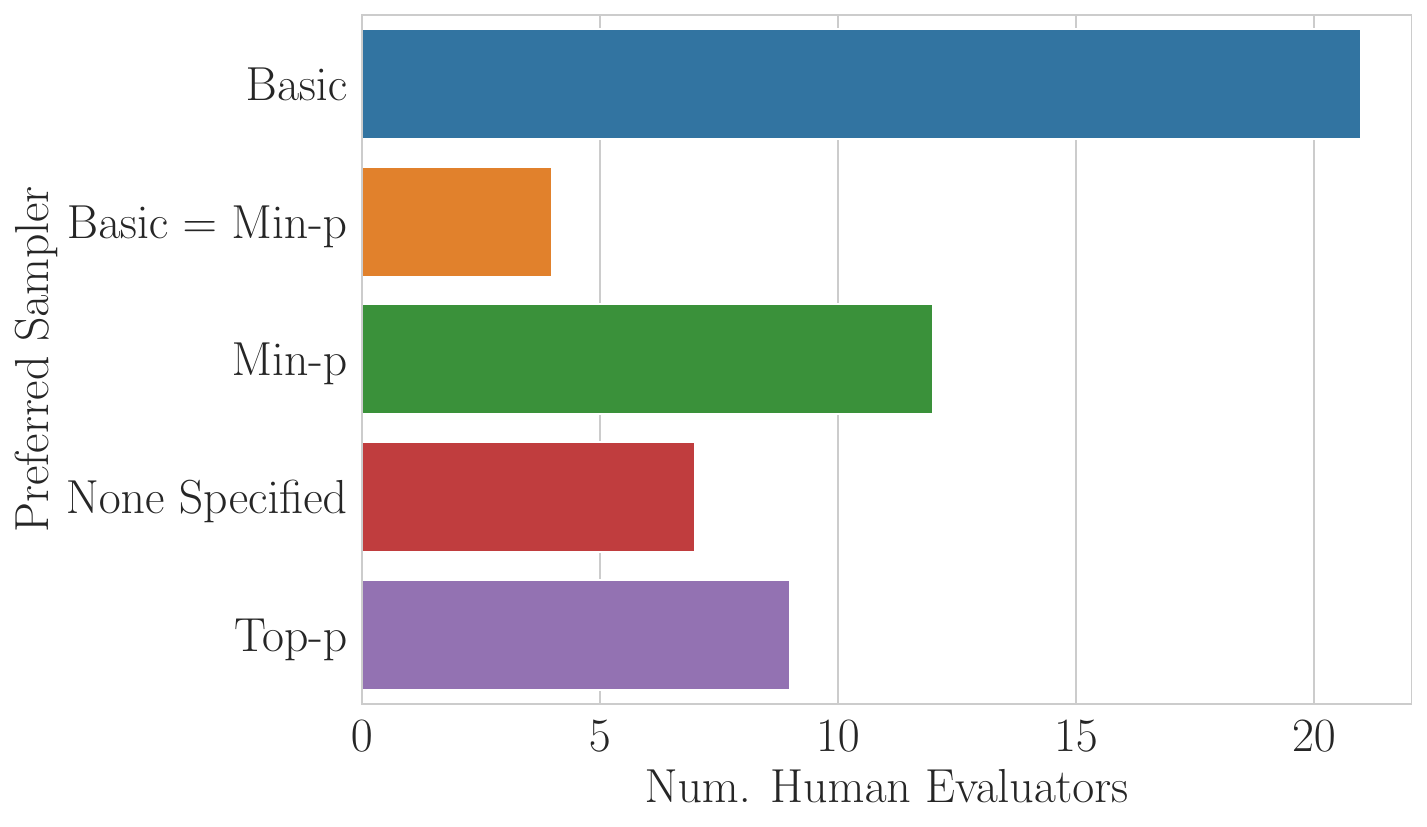}
    \caption{\textbf{Manual Annotation of Human Evaluators' Qualitative Responses Fail to Support Claim that \texttt{Min-P} Was the Preferred Sampler.} We manually annotated responses from human annotators regarding their preferred sampler(s) at the end of the original paper's study. The responses suggest \texttt{min-p} was not the most preferred sampler. We provide example responses in Sec.~\ref{sec:human_evals:subsec:human_feedback}.}
    \label{fig:attentive_human_annotators_preferred_samplers}
\end{figure}

At the end of the human evaluation study, the original paper asked human participants to qualitatively describe which sampler(s) they preferred. The paper claimed that human evaluators' qualitative responses support \texttt{min-p} over \texttt{top-p}:
\begin{quote}
    ``Participants frequently noted that outputs generated with min-p sampling were more coherent and creative, especially at higher temperatures.''
\end{quote}
However, when reading through the  \href{https://github.com/menhguin/minp_paper/blob/main/min_p%20user%20preference%20study%20v2.0%20(Responses)%20-%20Form%20responses%201%20(1)%20(2).csv}{paper's data}, we believe that \textbf{the qualitative responses suggest a different preference pattern}.
We manually annotated the qualitative responses and visualized our annotations of the humans' expressed preferences (Fig.~\ref{fig:attentive_human_annotators_preferred_samplers}), and \href{https://docs.google.com/spreadsheets/d/1-KNh4-dbrXBafb9iUQINqFBjDrZvkScPNbGH6IENecQ/edit?gid=1288914057#gid=1288914057}{publicly posted our annotations} in the same format as the original paper.
We found two results: (1) more human evaluators explicitly preferred \texttt{basic} sampling than preferred \texttt{min-p} sampling, and (2) \texttt{min-p} was only slightly preferred over \texttt{top-p}.
We provide quotations from human evaluators favoring \texttt{basic} sampling in Appendix~\ref{app:sec:human_qual_responses}.

\subsection{New Human Evaluation Study Shows \texttt{Min-p} Does Not Outperform Baselines in Quality, in Diversity, or in a Tradeoff Between Quality and Diversity}
\label{sec:human_evals:subset:new_exp}

\begin{figure}[t!]
    \centering
    \includegraphics[width=\linewidth]{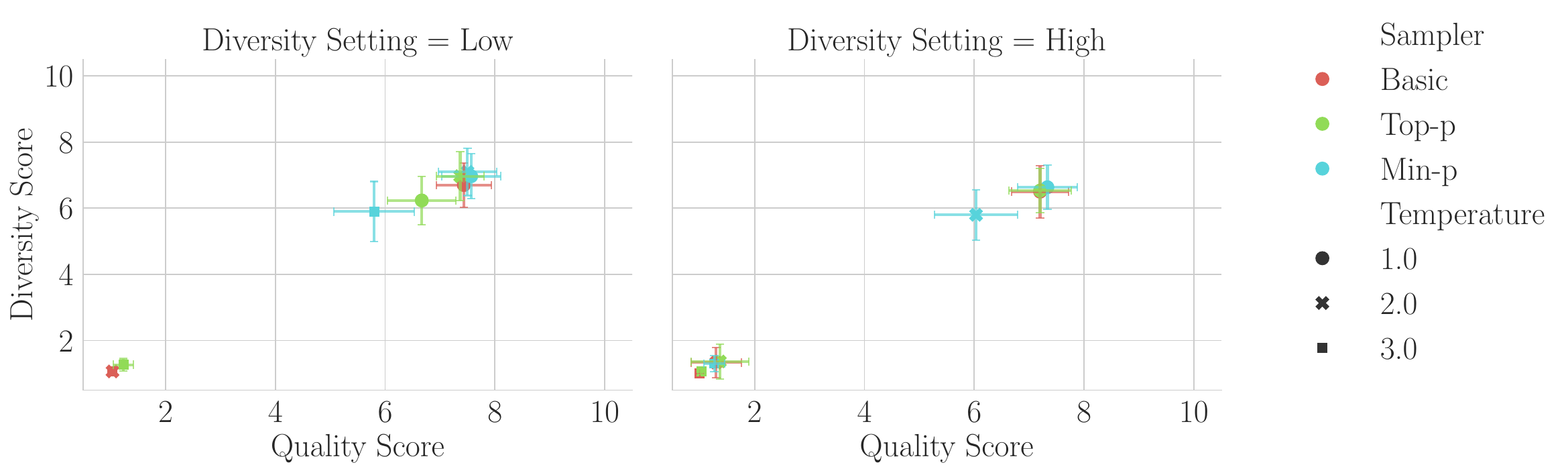}
    \captionof{figure}{\textbf{New Human Evaluation Study Suggests \texttt{Min-p} Does Not Outperform Baselines in Quality, in Diversity or in a Pareto-Optimal Tradeoff Between Quality and Diversity.} Visualization of scores from \citet{nguyen2024minp}'s second human experiment. \texttt{Min-p}'s performance advantage relative to \texttt{basic} and \texttt{top-p} sampling is observed in conditions (e.g., higher temperatures) where absolute quality and absolute diversity scores across all samplers are lower compared to other regimes (e.g., $T=1$). For practitioners optimizing for maximal quality and maximal diversity, these results suggest that \texttt{min-p} offers no apparent advantage over \texttt{basic} or \texttt{top-p} sampling.}
    \label{fig:new_human_evals}
\end{figure}

In response to our feedback, the authors conducted and added a new human evaluation study to Appendix C.2.
Their new study made multiple methodological changes:
\begin{itemize}
    \item Different sampler implementation: switched from applying temperature \textit{after} truncation to applying temperature \textit{before} truncation.
    \item Different distribution of human participants from Prolific.
    \item Different sampling hyperparameters for \texttt{top-p}: switched from $0.1$ and $0.9$ to $0.9$ and $0.95$.
    \item Different sampling hyperparameters for \texttt{min-p}: switched from $0.2$ and $0.05$ to $0.1$ and $0.05$.
    \item Different allotted reading time: increased from 30 minutes to 45 minutes.
    \item Different sampled text: 3 short paragraphs were replaced with a single complete story.
    \item Different rubric for human participants to evaluate sampled outputs.
\end{itemize}
Regarding the \href{https://github.com/menhguin/minp_paper/blob/main/%5BLIVE%5D%20min_p%20user%20preference%20study%20v3.0%20(Responses)%20-%20Form%20responses%201.csv}{new human evaluation data} and results, we share two discoveries here: First, we believe one value is incorrectly reported: in \citet{nguyen2024minp}'s Table 15, the average score of \texttt{min-p} at $p=0.05$ and temperature $T=2$ is reported as $7.80$, but based on the authors' publicly posted data, we believe the correct numerical value should be $5.80$. Second, more generally, the data show again that \texttt{Min-p} does not outperform baselines in quality, in diversity or in a favorable tradeoff between quality and diversity. In this new study, whenever \texttt{min-p} outperforms other samplers, it does so under conditions that yield lower absolute scores than other conditions (Fig.~\ref{fig:new_human_evals}). For instance, \texttt{min-p} shows an advantage over the baselines in the ``high" diversity setting at $T=2$ and in the ``low" diversity setting at $T=3$. However, in both of these conditions, \texttt{min-p} receives lower quality and diversity scores than it does in the ``high" diversity setting at $T=1$ and the ``low" diversity setting at $T=2$. This shows \texttt{min-p}'s advantage is observed primarily under conditions that yield lower overall quality and diversity scores compared to other achievable conditions. \textbf{For anyone seeking higher quality or diversity, \texttt{min-p} offers no apparent advantage over \texttt{basic} or \texttt{top-p} sampling}.

\section{Extending \texttt{Min-p's} NLP Benchmark Evaluations}
\label{sec:nlp_benchmark_evals}

We next turned to the original paper's NLP benchmark evaluations of several models on GSM8K with Chain-of-Thought \citep{cobbe2021gsm8k} and GPQA (5-shot) \citep{rein2023gpqa}, which concluded that:
\begin{quote}
``\texttt{Min-p} sampling achieves superior performance across benchmarks and temperatures.''
\end{quote}


\subsection{Thorough Hyperparameter Sweep on GSM8K Contradicts Claim of \texttt{Min-p}'s Superiority}

\begin{figure}[t!]
    \centering
    \includegraphics[width=0.8\linewidth]{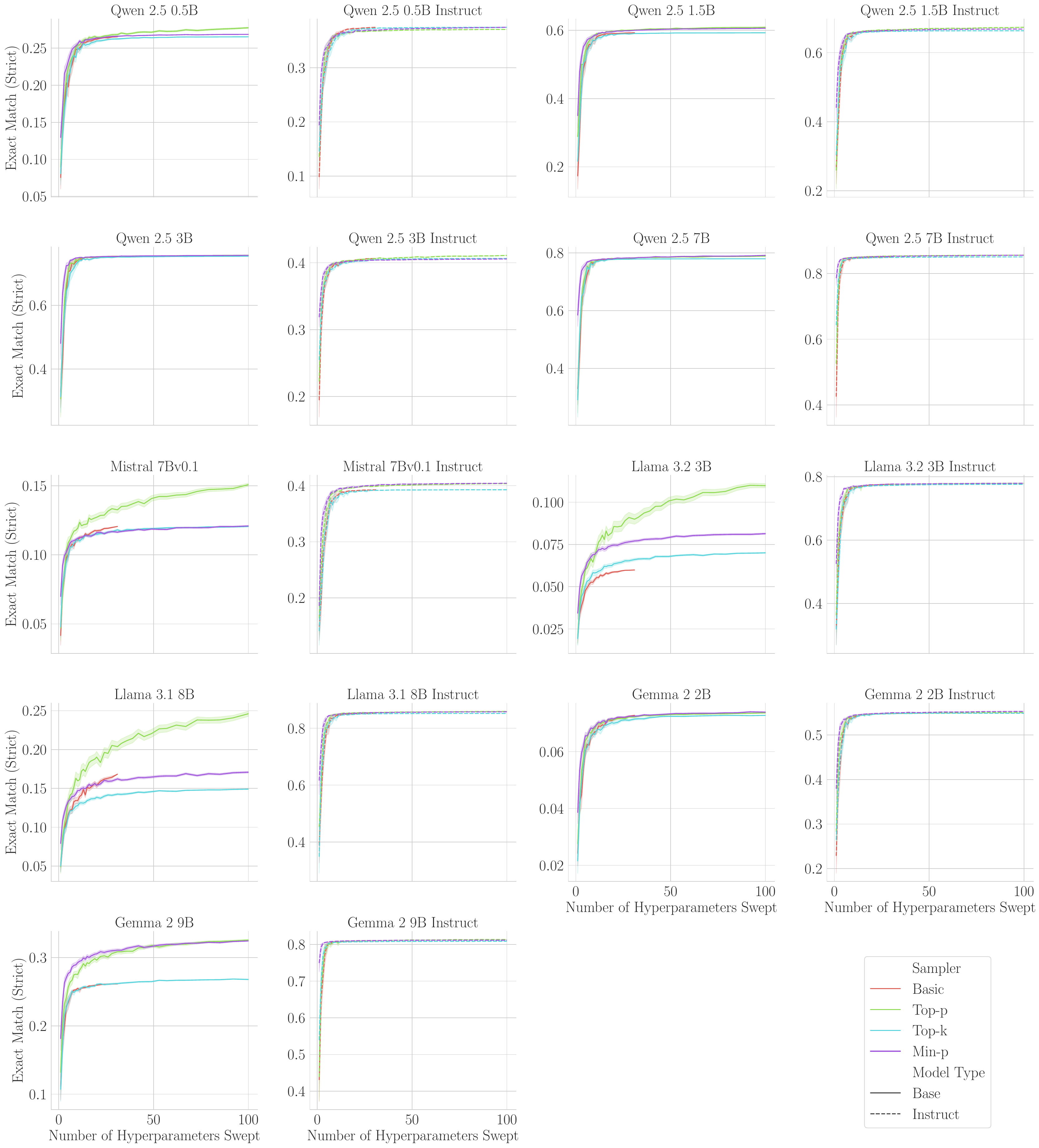}
    \caption{\textbf{\texttt{Min-P} Does Not Consistently Outperform Other Samplers on GSM8K When Controlling For Hyperparameter Volume.}
    In our first analysis, we measured how the maximum Exact Match (Strict) for each sampler improves as the number of hyperparameters increases. \texttt{Basic} sampling has only a temperature hyperparameter, and we therefore do not sweep it to the same degree.
    }
    \label{fig:gsm8k_em_strict}
\end{figure}

To test whether \texttt{min-p} indeed achieves superior performance, we conducted an extensive analysis on GSM8K CoT, sweeping the following models, samplers, hyperparameters and sampling seeds:
\begin{itemize}
    \item \textbf{9 Models:} Qwen 2.5 \citep{qwen2025qwen25technicalreport} 0.5B, 1.5B, 3B and 7B; Mistral 7Bv0.1 \citep{jiang2023mistral7b}; Llama \citep{grattafiori2024llama3herdmodels} 3.1 8B and 3.2 3B; Gemma 2 \citep{gemmateam2024gemma2improvingopen} 2B and 9B.
    \item \textbf{2 Model Stages:} Pre-trained (``Base") and Post-Trained (``Instruct").
    \item \textbf{4 Samplers:} \texttt{basic}, \texttt{top-p}, \texttt{top-k}, \texttt{min-p}.
    \item \textbf{31 Temperatures:} $0.0$ (``greedy") to $3.0$ in increments of $0.1$.
    \item \textbf{6 Hyperparameters Per Sampler:} We chose $6$ hyperparameters per sampler, except for \texttt{basic} which has no hyperparameter beyond temperature. The values were taken from the original paper; some were lightly edited to make them more evenly distributed:
    \begin{itemize}
        \item \texttt{basic}: No hyperparameters other than temperature.
        \item \texttt{top-k}:  $k \in \{10, 30, 50, 100, 150, 200 \}$.
        \item \texttt{top-p}: $p \in \{0.99, 0.98, 0.95, 0.9, 0.8, 0.7 \}$.
        \item \texttt{min-p}: $p \in \{0.01, 0.02, 0.05, 0.1, 0.2, 0.3 \}$.
    \end{itemize}
    \item \textbf{3 Random Seeds for Sampling:} $\{0, 1, 2\}$
\end{itemize}

\begin{figure}[t!]
    \centering
    \includegraphics[width=0.8\linewidth]{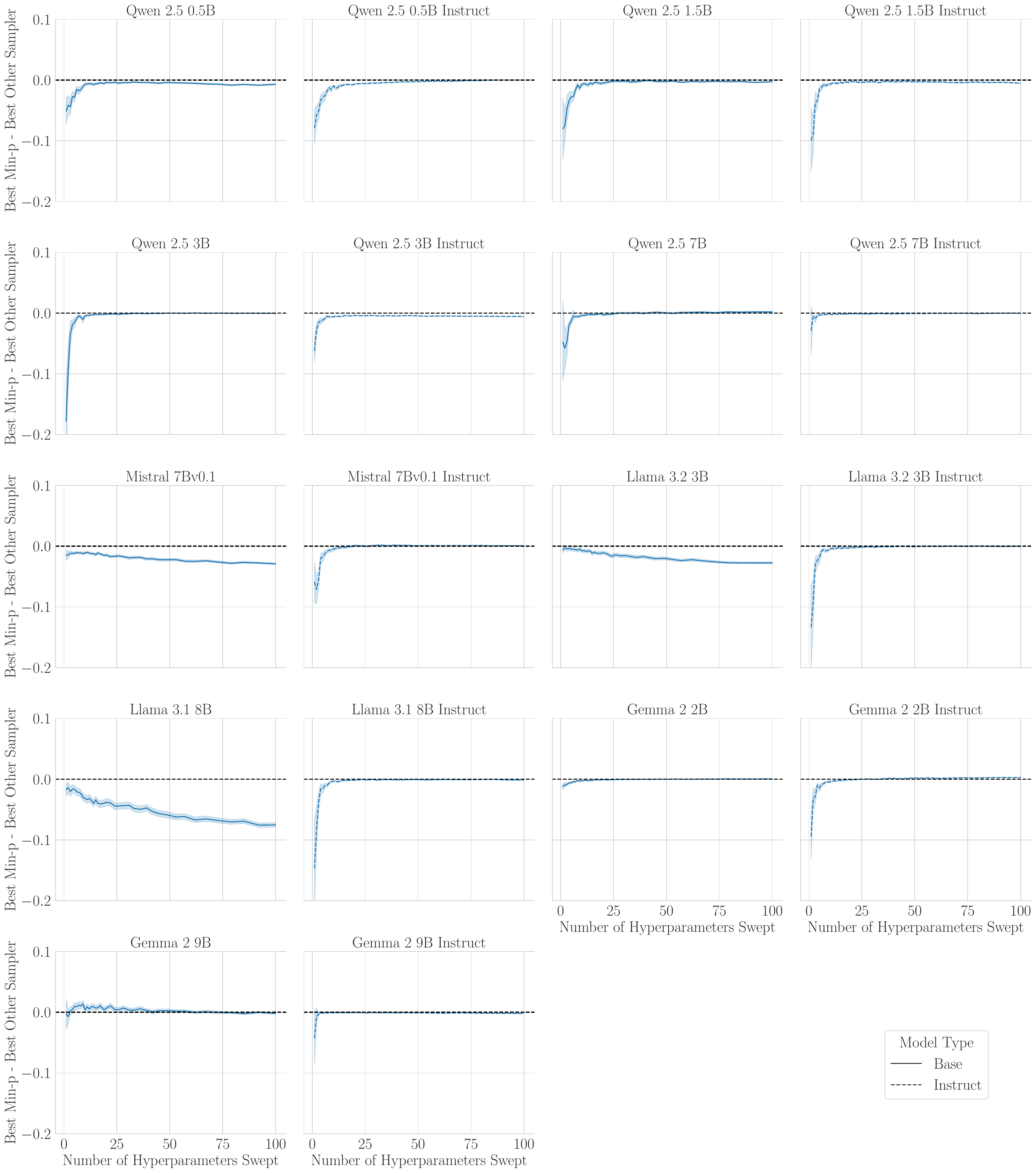}
    \caption{\textbf{\texttt{Min-P} Does Not Consistently Outperform Other Samplers on GSM8K When Controlling For Hyperparameter Volume.}
    In our second analysis, we measured how the difference between \texttt{min-p}'s highest score and the best non-\texttt{min-p} sampler's highest score changes as the number of swept hyperparameters increases. \texttt{Min-p} matches or underperforms other samplers.}
    \label{fig:gsm8k_diff_of_em_strict}
\end{figure}

Due to our compute budget, we only evaluated GSM8K CoT (albeit under two prompt formats, for reasons explained below). This sweep and the sweep below required $\sim6000$ Nvidia A100-hours.
GSM8K contains a subset of samples with ambiguous language or incorrect labels that have since been identified and cleaned \citep{vendrow2025large} and that models may have been trained on GSM8K \citep{zhang2024careful}, but we used GSM8K nonetheless for consistency with the original paper. We similarly used EleutherAI's LM Eval Harness \citep{gao2021framework,biderman2024lessonstrenchesreproducibleevaluation}. 

To evaluate how performant each sampler is, we first averaged over the three sampling seeds and then conducted two complementary analyses:
\begin{enumerate}
    \item For each sampler, we subsampled an equal number of hyperparameters ranging from $N=1$ to $N=100$ and computed the maximum Exact Match (Strict) score achieved by the sampled subset of size $N$. We repeated this process $150$ times, averaging over the subsampled subsets' scores. This ``Best-of-N" analysis \citep{nakano2021webgpt,stiennon2020learning,hughes2024bestofnjailbreaking,schaeffer2025largelanguagemonkeyspower} tells us the best possible performance each sampler will likely obtain as its hyperparameter space increases.
    
    \item For $N=1$ to $N=100$, we subsampled $N$ hyperparameters per sampler and computed the difference of the maximum Exact Match (Strict) score achieved by \texttt{min-p} minus the maximum score achieved by any other sampler. We repeated this process $150$ times, averaging over the subsampled subsets. This tells us by how much \texttt{min-p} outperforms all other samplers, controlling for the size of hyperparameter space of each sampler.
\end{enumerate}

\textbf{Both analyses reached consistent results: \texttt{min-p} does not outperform other samplers when equalizing the volume of hyperparameter space.}
Fig.~\ref{fig:gsm8k_em_strict} and Fig.~\ref{fig:gsm8k_diff_of_em_strict} respectively demonstrate that min-p is largely indistinguishable from other samplers.

After we showed these results to the authors, they informed us that we had run our experiments using the \href{https://github.com/EleutherAI/lm-evaluation-harness/blob/main/lm_eval/tasks/gsm8k/gsm8k-cot-llama.yaml}{``Llama"
formatting of GSM8K CoT prompts} as we used the command from the authors'
\href{https://github.com/menhguin/minp_paper/blob/main/%5BPUBLIC%5D_Min_P_Evals_Replication_for_GPQA_and_GSM8K_COT.ipynb}{public Colab notebook};
the authors \href{https://github.com/menhguin/minp_paper/commit/9ca63687df807b4276c527b9b47e65379c1b0bd6}{clarified} that "Llama" formatting should be used only for Llama models.
We then reran our experiments using \href{https://github.com/EleutherAI/lm-evaluation-harness/blob/main/lm_eval/tasks/gsm8k/gsm8k-cot.yaml}{standard formatting of GSM8K CoT prompts}.
The results were nearly identical (Appendix \ref{app:sec:gsm_cot_scores}), with one small difference: \texttt{min-p} does produce higher scores for 2 of 12 language models.
Again, we conclude \textbf{\texttt{min-p} does not outperform other samplers on either formatting of GSM8K CoT when controlling for hyperparameter volume}.

\section{Investigating \texttt{Min-p}'s LLM-As-A-Judge Evaluations}
\label{sec:llm_as_judge_evals}

We then turned to the original paper's LLM-as-a-Judge evaluations \citep{zheng2023judging}, specifically AlpacaEval creative writing evaluations \citep{dubois2023alpacafarm}.

\subsection{Under-Specified and Indirect Methodology Hinders Reproduction and Interpretation}

\begin{figure}[t!]
    \centering
    \includegraphics[width=0.9\linewidth]{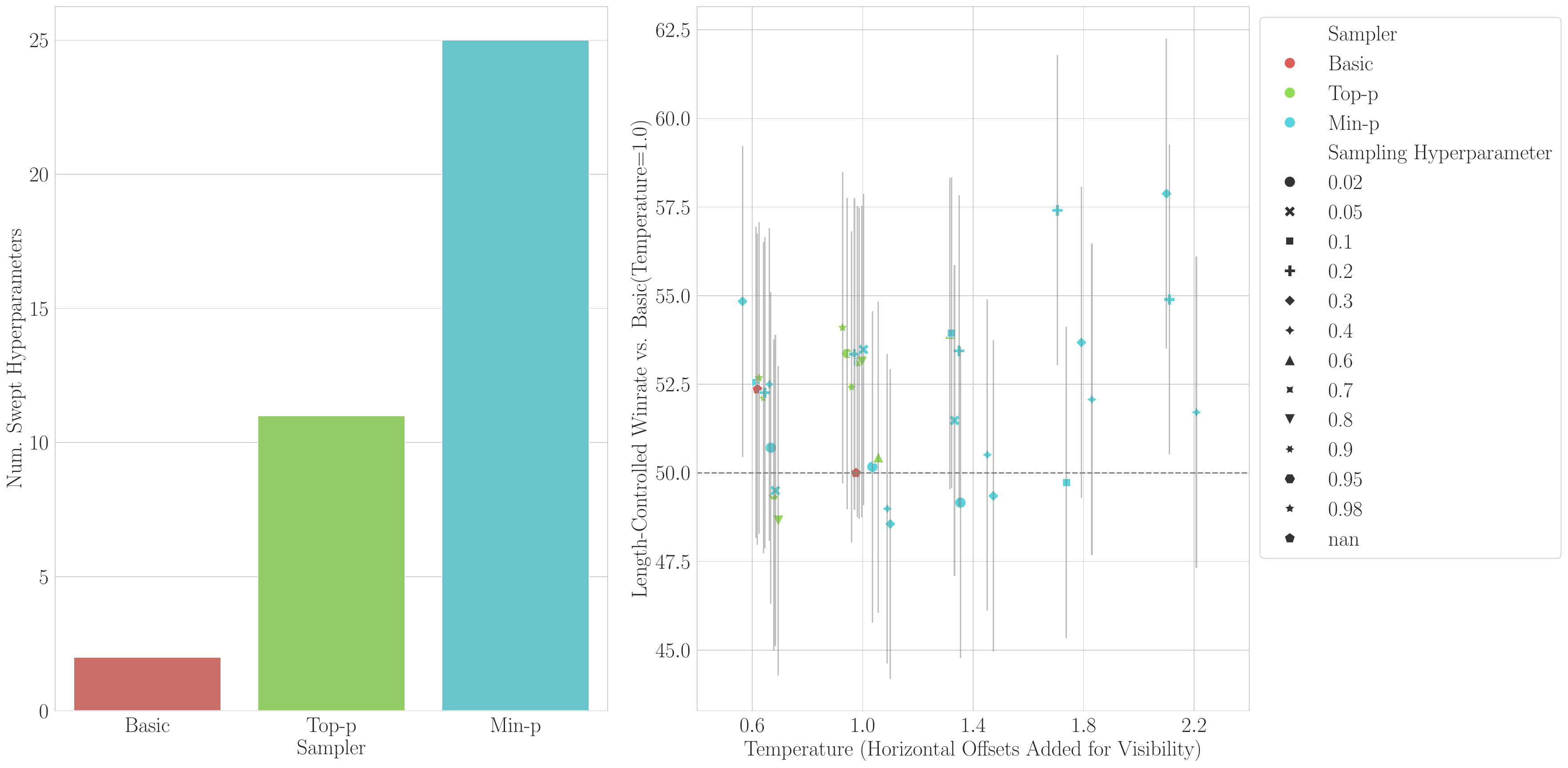}
    \caption{\textbf{\citet{nguyen2024minp}'s LLM-As-A-Judge Evaluations Suggest \texttt{Min-p} Typically Matches Other Samplers Despite $2\times$ to $10\times$ More Hyperparameter Tuning.} Left: \citet{nguyen2024minp} swept \texttt{min-p} with more than twice as many hyperparameters as \texttt{top-p} and more than ten times as many hyperparameters as \texttt{basic}. Right: Pairwise comparisons show \texttt{min-p} typically performs on-par with other samplers. Data were obtained from the first author's \href{https://github.com/menhguin/quest?tab=readme-ov-file}{public GitHub repository}.
    }
    \label{fig:alpacaeval}
\end{figure}

In the \href{https://arxiv.org/abs/2407.01082v2}{Oct 2024 Arxiv manuscript} and \href{https://openreview.net/notes/edits/attachment?id=SR6ORZeBAn&name=pdf}{ICLR OpenReview manuscript}, the methodology is under-specified in several ways: There is no mention which model(s) were sampled from, which model(s) served as the judge(s), or how hyperparameters were chosen or swept. Additionally, there is no description of uncertainty for the reported win rates, meaning readers are unable to decide whether win rates are statistically different from chance (50.00\%).

Furthermore, the experiment seems designed in a manner that introduces a confounder. For those unfamiliar, AlpacaEval reports win rates between paired comparisons.
Instead of directly comparing \texttt{min-p} against other samplers, the authors compared each sampler against a common fixed sampler: \texttt{basic}$(\tau=1.0)$.
This comparison strategy is indirect since comparing directly against \texttt{min-p} would offer a clearer test of its superiority while using the same number of comparisons.
The authors’ design choice is additionally concerning because LLM-judge preferences are probably not transitive, as shown by recent research \citep{xu2025investigatingnontransitivityllmasajudge}; that is, if sampler A beats sampler B, and sampler B beats sampler C, it does not necessarily follow that sampler A beats sampler C.
Therefore, comparing all methods to \texttt{basic}$(\tau=1.0)$ provides no reliable inference about \texttt{min-p}’s performance relative to \texttt{top-p} or \texttt{basic} at other temperatures. \textbf{These under-specified aspects of the methodology, combined with its indirect experimental design, make drawing conclusions difficult.}

\subsection{\texttt{Min-p} Received More Hyperparameter Tuning and Frequently Fails to Win}

No scores or code to create scores were provided with the original paper's \href{https://github.com/menhguin/minp_paper/issues/5}{GitHub repository}. While drafting this manuscript, we became aware of \href{https://github.com/IlyaGusev/quest/compare/main...menhguin:quest:main}{ongoing work} to release code in a separate repository. Results from that code revealed two discoveries: First, \texttt{min-p} received $\sim2\times$ more hyperparameter tuning than \texttt{top-p} sampling and $\sim 10\times$ more tuning than \texttt{basic} sampling (Fig.~\ref{fig:alpacaeval}, left), potentially tilting the scales in its favor. Second, the win-rates show that \texttt{min-p} frequently fails to outperform \texttt{top-p} and \texttt{basic} sampling, especially when accounting for confidence intervals; we visualized the new data with 95\% confidence intervals (with horizontal offsets added for visibility) (Fig.~\ref{fig:alpacaeval}, right).

\subsection{Table 3(b) Reported The Higher of Two Scores For \texttt{Min-p} But the Lower of Two Scores For \texttt{Top-p}}

As evidence for the LLM-As-A-Judge evaluation scores in the original paper's Table 3(b), the first author \href{https://t.me/senior_augur/76}{publicly shared a Telegram link} that showed the higher of two scores was reported for \texttt{min-p} (the reported win rate of $52.01$ corresponds to $p=0.05$, but $p=0.01$ yields a lower win rate of $50.14$) but the lower of two score was reported for \texttt{top-p} (the reported win rate of $50.07$ corresponds to $p=0.9$, but $p=0.98$ yields a higher win rate of $50.43$).
%

\section{Substantiating \texttt{Min-p}'s Community Adoption Claims}
\label{sec:community_adoption}

\subsection{Claimed GitHub Repositories \& Stars Were Unsubstantiated and Retracted}

The Arxiv and peer-reviewed manuscripts of \citet{nguyen2024minp} included specific claims about \texttt{min-p}'s adoption in the language modeling community:
\begin{quote}
    ``\textbf{Community Adoption:} Min-p sampling has been rapidly adopted by the open-source community, with over 54,000 GitHub repositories using it, amassing a cumulative 1.1 million stars across these projects."
\end{quote}
We attempted to verify these numbers through analysis of major GitHub language modeling repositories.
Per our calculations, the combined GitHub stars of leading LM repositories (\texttt{transformers}, \texttt{ollama}, \texttt{llama.cpp}, \texttt{vLLM}, \texttt{Unsloth}, \texttt{mamba}, \texttt{SGLang}, \texttt{llama-cpp-python}) sum to 453k stars as of March 2025, less than half the 1.1M stars claimed by \texttt{min-p} alone.
We could not substantiate either 49k GitHub repositories or 1.1M GitHub stars. 
When we inquired how these numbers were calculated, the authors \href{https://github.com/menhguin/minp_paper/issues/6#issuecomment-2686274162}{publicly stated that GitHub was searched for ``min-p''}, which yields many false positives. \textbf{The authors retracted both the 54k GitHub repository claim and the 1.1M GitHub stars claim from the ICLR 2025 Camera Ready manuscript.}

Given that the numbers have been retracted, we debated whether to include this section.
We decided to include it for three reasons.
First, we wanted to document this clear failure of the review process.
These numbers were unsubstantiated in the manuscript, and, in our opinion, were preposterous. 
Yet three out of four reviewers and the Area Chair highly commended the community adoption as evidence of \texttt{min-p}'s superiority;
for instance, the \href{https://openreview.net/forum?id=FBkpCyujtS&noteId=WH4Y2gZs6R}{Area Chair wrote}:
\begin{quote}
    ``[\texttt{Min-p}] is simple and is already widely adopted by the community (as mentioned by [Reviewer] D38H, “The usage of it in 54,000 Github repositories alone is very impressive”). [...] The resulting review scores reflect the high quality of the paper: It presents convincing experiments, thorough analysis, and the provided method has an extremely high impact."
\end{quote}
\href{https://openreview.net/forum?id=FBkpCyujtS&noteId=tMSeNlMmSa}{Reviewer fwNb similarly emphasized} the community adoption numbers:
\begin{quote}
    ``[\texttt{min-p}] has good empirical results, both as measured on benchmarks and (more important [sic]) by adoption of the community"
\end{quote}

Second, the machine learning research community may have learned of \texttt{min-p} before these community adoption numbers were retracted, e.g., when the original paper was posted to ArXiV or accepted at ICLR 2025.
Thus, we felt a proactive clarification would better rectify the scientific record.
Third, as we detail below, we believe the new community adoption statement remains misleading.

\subsection{The Revised Community Adoption Statement Inflates \texttt{Min-p}’s Adoption}

The ICLR 2025 Camera Ready now has a different statement of community adoption:
\begin{quote}
    ``[\texttt{Min-p}] is now integrated in widely used frameworks such as Hugging Face Transformers, vLLM, and SGLang, which collectively have accrued over 350,000 GitHub stars. This integration, coupled with extensive downstream usage (e.g., over 290,000 dependent repositories for Transformers alone), underscores the method’s practical impact."
\end{quote}

While being integrated into such frameworks is indeed a contribution, this statement misleadingly represents these frameworks’ usage as \texttt{min-p}’s usage, rather than specifically measuring \texttt{min-p}’s usage.
This new statement is akin to publishing a book and then claiming credit for the library.

\section{Discussion and Limitations}
\label{sec:discussion}

\paragraph{Scientific Conclusions} This investigation led us to conclude that the four lines of evidence presented by \citet{nguyen2024minp} -- (1) human evaluations, (2) NLP benchmark evaluations, (3) LLM-as-a-Judge evaluations, (4) community adoption -- do not support claims of \texttt{min-p}'s superiority.
While \texttt{min-p} is useful for providing users another options to try, the original paper's data and our extensions of the original paper's data suggest that all samplers perform roughly the same once given the same amount of hyperparameter tuning; however, in our view, more research would be needed to assess the veracity of this conclusion.
The paper's data does weakly suggest that \texttt{min-p} sampling can sometimes provide a benefit at higher temperatures, albeit with the critical caveat that absolute performance is meaningfully worse in this high-temperature regime than in standard temperature regimes.

\paragraph{Key Limitation} Our manuscript re-analyzes the evidence presented by the original paper \citep{nguyen2024minp} and additional evidence created using the original paper's code. \textit{Conclusions here are based on that evidence}. We emphasize that new evidence might lead to different conclusions.

\paragraph{What Went Wrong During the ICLR 2025 Review Process?}

\citet{nguyen2024minp}'s outstanding success in the ICLR 2025 review process---achieving Oral presentation status and ranking as the 18th highest-scoring submission\footnote{\url{https://papercopilot.com/statistics/iclr-statistics/iclr-2025-statistics/}}---is difficult to reconcile with the flaws our investigation uncovered.

The reviewers overlooked methodological issues such as which model(s) are being sampled from
for the LLM-as-judge evals and missing/inadequate/improper consideration of uncertainty in presented results.
Reviewers uncritically accepted the authors' claim that "over 54,000 GitHub repositories" were using \texttt{min-p} sampling, 
when intuition or a quick GitHub search reveals cause for pause.

The Area Chair's comment also contains a clear misstatement: it highlights \texttt{min-p}'s success in the low temperature regime ("in the low temperature regime, [\texttt{min-p}] provides a significant advantage"), when the paper specifically claims benefits in the \emph{high} temperature regime.
\clearpage

\bibliography{references_rylan}
\bibliographystyle{colm2025_conference}

\clearpage

\appendix

\section{Examples of Human Qualitative Responses Favoring \texttt{Basic} Sampling Over \texttt{Min-P} Sampling}
\label{app:sec:human_qual_responses}

In Section~\ref{sec:human_evals:subsec:human_feedback}, we described how qualitative responses from many human participants in the original paper's study favored \texttt{basic} sampling. Direct quotes from human evaluators favoring \texttt{basic} sampling are provided below. In the study, \texttt{basic} sampling was called “Model A”; for clarity, we substituted the pseudonyms for the actual sampling methods):
\begin{itemize}
    \item ``[\texttt{basic} sampling] on Temp 3.0 - High Diversity setting. The stories where [sic] more interenting [sic], felt more different compared to the others, which felt like the same ideia [sic] just in a different format.”
    \item ``I felt like [\texttt{basic} sampling] was most diverse and most interesting with it's [sic] descriptions of the characters and the setting. It appealed to me most and seemed to have less 'broken' sentences that didn't make sense. Descriptions were painterly [sic] and elaborate.”
    \item ``[\texttt{basic} sampling] was more engaging, it aroused my curiosity.”
    \item ``[\texttt{basic} sampling] provided more depth and easy to read for me and there was more diversity.”
    \item ``[\texttt{basic} sampling], they presented creative storytelling”
    \item ``[\texttt{basic} sampling]. From the very beginning the verbiage and descriptions were very creative and vivid. And each story was unique”
    \item ``I believe that [\texttt{basic} sampling] has provided stories with more differentiation overall than the other two models. From the point of view of creativity, all three models are more or less equivalent as they almost always talk about stories set in extraterrestrial worlds both from a physical and mental (dreams) point of view"
    \item ``[\texttt{Basic} sampling]: Sample 2: Temperature Setting F (Temp 3.0 - High Diversity). The story was captivating, it took inside the mystical land and walked you right besides all the characters, you can even draw the characters from just th descriptions provided by the prompt. you Could even smell them, smell the setting and be at one with the setting."
    \item ``I personally preferred [\texttt{basic} sampling] on the setting of creative, descriptive storytelling. I enjoyed how the writing was creative, showing imagination and a strong use of language. The stories were quite evocative, with intriguing settings and characters that helped to draw the reader in. I also appreciated the diversity of themes that were explored, from night weavers to dream manipulation and mysterious libraries, which kept the stories engaging and interesting."
    \item ``Temporature setting C on [\texttt{basic} sampling] was the best. The story was fascinating and very engaging. I wanted to read more."
    \item ``I prefered the first [\texttt{basic} sampling]. Tho [\texttt{basic} sampling] and C seem to be very head to head. But something about [\texttt{basic} sampling] seemed different in quality about it to me."
\end{itemize}

More quotes are in the \href{https://github.com/menhguin/minp_paper/blob/main/min_p%20user%20preference%20study%20v2.0%20(Responses)%20-%20Form%20responses%201%20(1)%20(2).csv)}{original paper's data}. We urge readers to draw their own conclusions.

\clearpage
\section{GSM8K Chain-of-Thought Scores with ``Standard" Formatting}
\label{app:sec:gsm_cot_scores}

At the request of \citet{nguyen2024minp}, we reran our GSM8K Chain-of-Thought sweeps using ``standard" formatting instead of ``Llama" formatting. \textbf{Both analyses reached consistent results: \texttt{min-p} does not consistently outperform other samplers when controlling the volume of hyperparameter space.}

\begin{figure}[b!]
    \centering
    \includegraphics[width=1.0\linewidth]{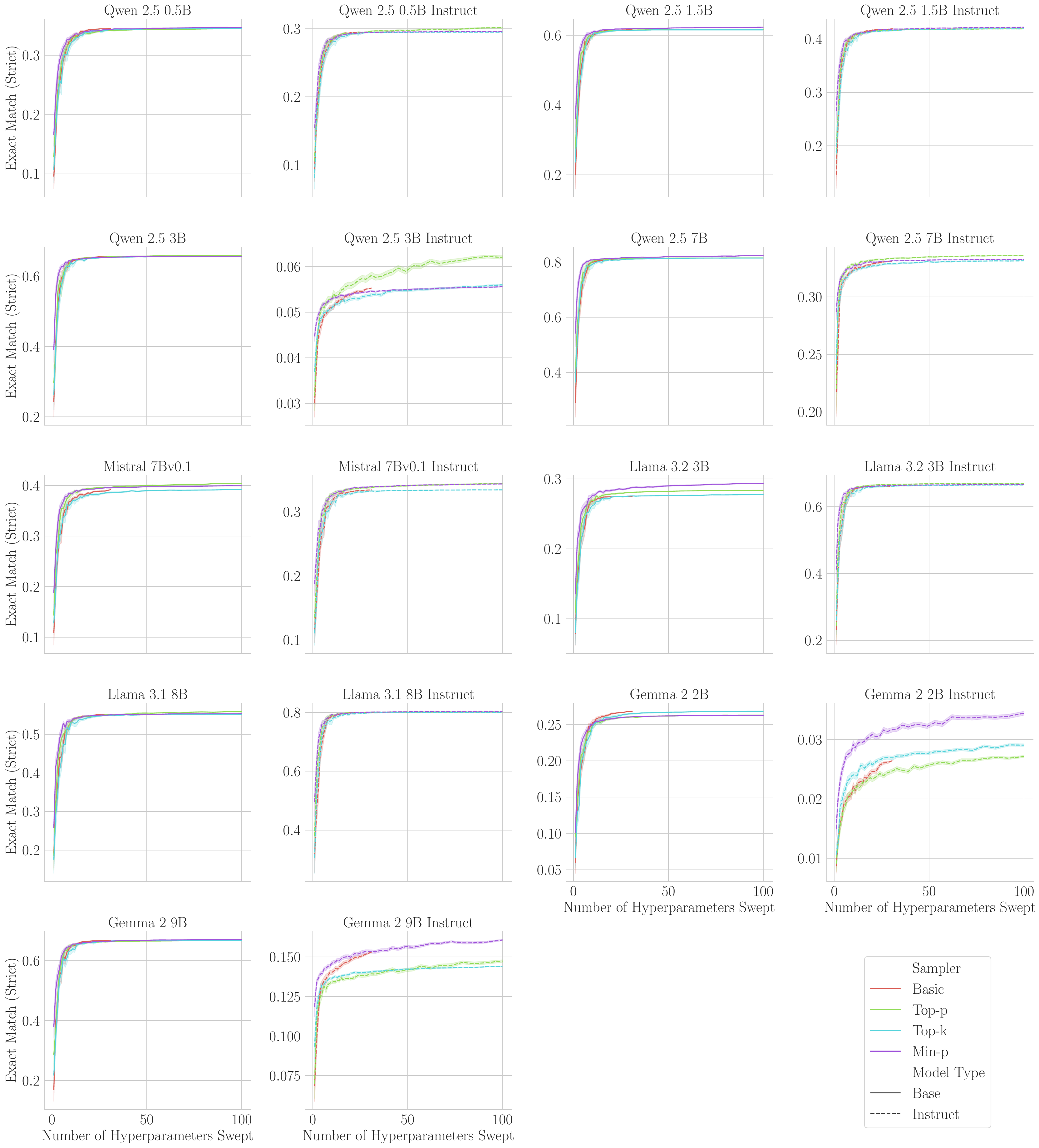}
    \caption{\textbf{\texttt{Min-P} Does Not Consistently Outperform Other Samplers on GSM8K When Controlling For Hyperparameter Volume.}
    We reran our GSM8K sweep using ``standard" formatting rather than ``Llama" formatting and observed qualitatively similar data.
    }
\end{figure}

\clearpage

\begin{figure}[t!]
    \centering
    \includegraphics[width=1.0\linewidth]{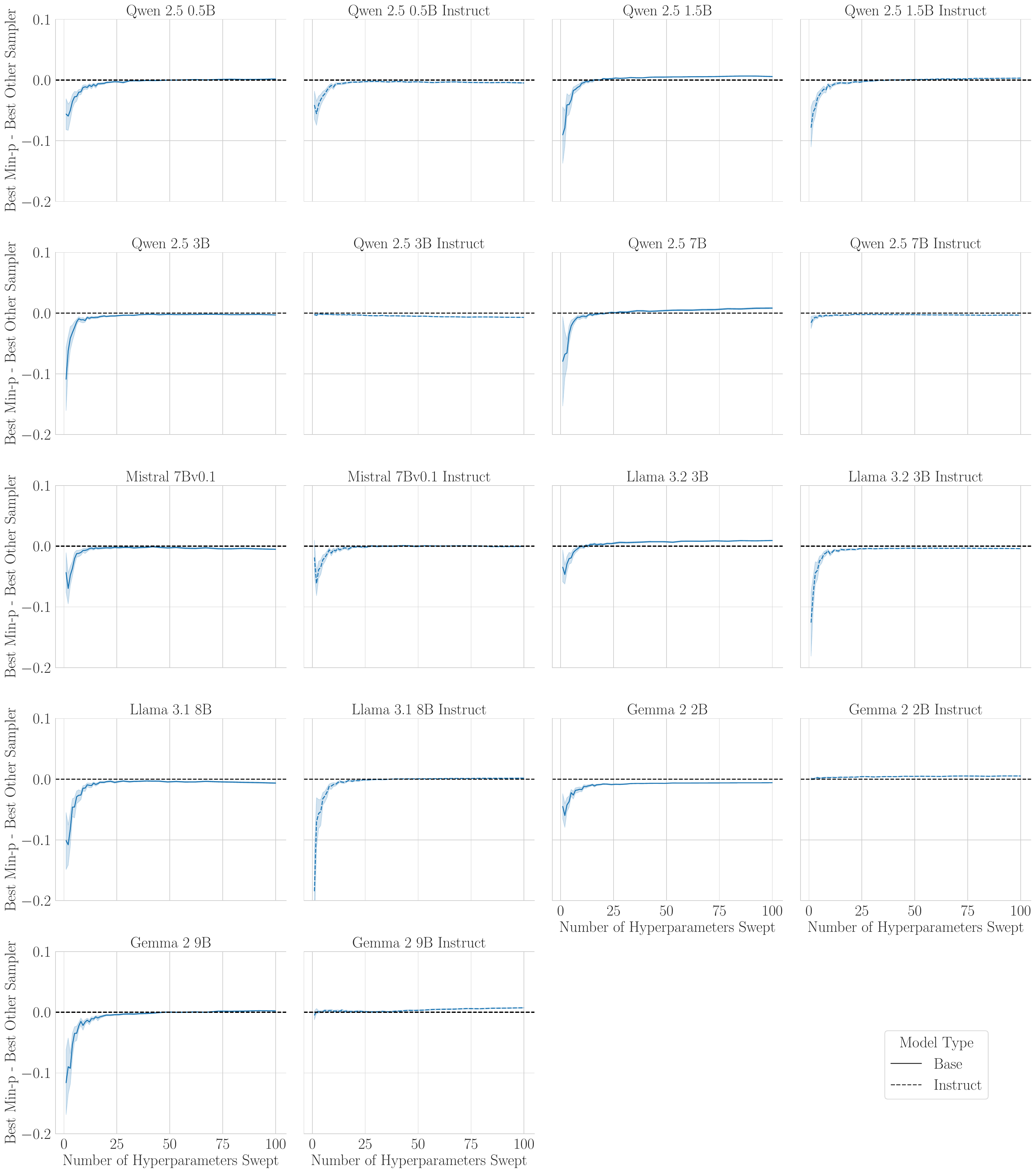}
    \caption{\textbf{\texttt{Min-P} Does Not Consistently Outperform Other Samplers on GSM8K When Controlling For Hyperparameter Volume.}
    We reran our GSM8K sweep using ``standard" formatting rather than ``Llama" formatting and observed qualitatively similar data.}
\end{figure}

\clearpage

\clearpage

\section{GSM8K Scores By Model, Sampler and Hyperparameters}

\begin{figure}[b!]
    \centering
    \includegraphics[width=0.41\linewidth]{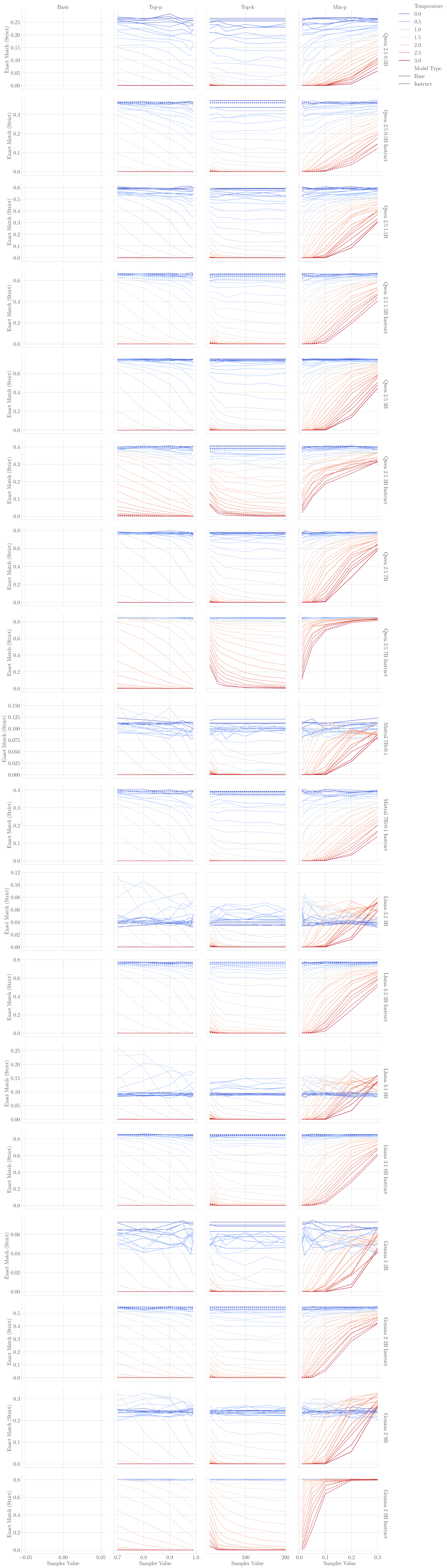}
    \caption{\textbf{GSM8K Scores By Model, Sampler and Sampler Hyperparameters.} Many models achieve their highest scores at low temperatures across samplers.}
    \label{fig:gsm8k_per_sampler_scores}
\end{figure}

\end{document}